\documentclass[letterpaper, 10 pt, conference]{IEEEtran}  

\usepackage{lmodern}
\usepackage[T1]{fontenc}

\IEEEoverridecommandlockouts                              
\usepackage{graphics} 
\usepackage{epsfig} 
\usepackage{mathptmx} 
\usepackage{times}
\usepackage{amsmath} 
\usepackage{amssymb} 
\usepackage{xcolor} 
\usepackage{subcaption}
\usepackage{wrapfig}
\usepackage{algorithm}
\usepackage{algpseudocode}
\usepackage{url}
\usepackage{caption}
\usepackage{subcaption}
\usepackage{epstopdf}
\usepackage{multicol}
\usepackage{multirow}
\usepackage{times}

\title{\LARGE \bf
Learning Approach to Efficient Vision-based Active Tracking of a Flying Target by an Unmanned Aerial Vehicle
}

\author{Jagadeswara PKV Pothuri$^{1}$, Aditya Bhatt$^{1}$, Prajit KrisshnaKumar$^{1}$, Manaswin Oddiraju$^{1}$ and Souma Chowdhury$^\dagger$ 
\thanks{$^\dagger$ Corresponding Author, soumacho@buffalo.edu}
\thanks{$^{1}$ Mechanical \& Aerospace Eng.,
        University at Buffalo, Buffalo, NY 
        }%
\thanks{Copyright \textcopyright 2025 by Souma Chowdhury. Published by the American Institute of Aeronautics and Astronautics, Inc., with permission}
}

\begin{document}

\maketitle

\begin{abstract}
Autonomous tracking of flying aerial objects has important civilian and defense applications, ranging from search and rescue to counter-unmanned aerial systems (counter-UAS). Ground based tracking requires setting up infrastructure, could be range limited, and may not be feasible in remote areas, crowded cities or in dense vegetation areas. Vision based active tracking of aerial objects from another airborne vehicle, e.g., a chaser unmanned aerial vehicle (UAV), promises to fill this important gap, along with serving aerial coordination use cases. Vision-based active tracking by a UAV entails solving two coupled problems: 1) compute-efficient and accurate (target) object detection and target state estimation; and 2) maneuver decisions to ensure that the target remains in the field of view in the future time-steps and favorably positioned for continued detection. As a solution to the first problem, this paper presents a novel integration of standard deep learning based architectures with Kernelized Correlation Filter (KCF) to achieve compute-efficient object detection without compromising accuracy, unlike standalone learning or filtering approaches. The proposed perception framework is validated using a lab-scale setup. For the second problem, to obviate the linearity assumptions and background variations limiting effectiveness of the traditional controllers, we present the use of reinforcement learning to train a neuro-controller for fast computation of velocity maneuvers. New state space, action space and reward formulations are developed for this purpose, and training is performed in simulation using AirSim. The trained model is also tested in AirSim with respect to complex target maneuvers, and is found to outperform a baseline PID control in terms of tracking up-time and average distance maintained (from the target) during tracking. 



\end{abstract}

\section{Introduction}

Vision based air-to-air tracking of an aerial object is an important and challenging problem with potential applications spanning collision avoidance, multi-aircraft coordination, and counter unmanned aerial system (counter-UAS) scenarios \cite{wang2021counter, castrillo2022review}. In such applications, the ability to track the flying object from another autonomous aircraft or uncrewed aerial vehicle (UAV) -- the chaser -- would expand operational capabilities, in terms of area, speed and agility of tracking and enable quickness of response that might otherwise be difficult with ground based tracking \cite{poitevin2017challenges}. 
In this paper, we premise that the success of such vision-based air-to-air tracking depends on achieving the following goals: 
\textbf{1)} efficient target detection,
\textbf{2)} efficient estimation of the target's motion, and  
\textbf{3)} efficient control of the chaser to support continued detection.
In addition to these goals, identifying the intent of the target flying object, could further improve the tracking success \cite{sun2023deep}; however, intent prediction is not within the scope of this paper. 

The deemed (speed, remote area coverage, agility) advantages of air-to-air tracking necessitates the decision-making to happen onboard, since the communication range, quality and latency required to support data processing and interpretation at a remote station would be impractical in most scenarios. 
With on-board computation, the computational complexity of the algorithm and the target speed (given that its trajectory is not known a priori)
pose critical challenges to successful tracking. 
Tracking a target with high acceleration usually requires processing more frames per second (FPS). 
For example, consider the relative target speed of 5 m/s. If the visual tracking algorithm allows 5 FPS versus 20 FPS, it means that the target will have moved 1 m versus 0.25 m in two consecutive frames captured by the chaser. 
Thus, the former case leads to more uncertainties in the target state predictions and hence makes it difficult for the control algorithm to give optimal control inputs to the chaser.
In the cases where tracking has to be performed by a UAV, the challenges are even more pronounced since the compute resources are constrained by the size and battery capacity of the UAV. 
In this paper, we present a robust yet computationally efficient framework for real-time tracking of a flying target. 

The active tracking problem in general is decomposed into two sub-problems: detection/estimation of the target object and control of the own vehicle \cite{sattigeri2007vision, li2024application, kainth2023chasing, yamauchi2020visual}. We adopt this decomposition concept in our work here as well. In the remainder of this introductory section, we motivate our approach by discussing the capabilities and limitations of the existing frameworks to solve each of these sub-problems. Thereafter we converge on the objectives and key contributions of this paper.  


With regards to target detection, there exist several frameworks in the computer vision domains that essentially track an object in an image. Tracking an object in an image will now be referred as visual tracking and the active tracking corresponds to the visual tracking of the target along with the chaser control. 
Correlation based visual tracking algorithms \cite{liu2021overview} have been touted to provide high computational efficiency such as with MOSSE \cite{bolme2010visual} (allows upto 600 FPS) and Kernelized Correlation Filter (KCF) \cite{henriques2012exploiting} (allows up to 290 FPS). However, these algorithms are usually tested on a video or a sequence of images generated from a static observer. Moreover, their accuracy suffers when there is a significant change in the background. Rapid changes in the background are expected in air-to-air tracking applications given that the target and the chaser both have 6 Degrees Of Freedom (DOF). These algorithms also work under the assumption that the target is in the field of view. In other words, the absence of target in a frame (when that occurs) cannot be inferred using these algorithms. On the other hand, there have been significant strides in developing deep learning based approaches that address these short-comings; such architectures includes those using convolutional neural networks (CNN) and Vision Transformer architectures (ViT) for object detection \cite{jiang2022review}. However, these approaches are usually data hungry in training and are often quite large in size in terms of number of parameters. These issues limit generalizability to  variety of scenarios and deployment onboard. In contrast, here we propose an algorithmic framework that seeks to jointly leverage the capabilities of correlation filter based approaches and deep-learning approaches to achieve suitable accuracy while maintaining acceptable computational efficiency (FPS) needed for active tracking applications. More specifically, in our work the visual tracking is performed by KCF, and then the target velocity estimations and a confidence measure (APCE) is used to invoke the CNN model YOLO (You Only Look Once). 
The standard visual tracking algorithms are usually validated on a benchmark video dataset \cite{fan2019lasot, hamdi2020drotrack}. Such video datasets are often recorded from a static observer and hence they may not suitable in evaluating the target state estimation for
active tracking problem where both the target and the camera  continuously get velocity inputs. 
Therefore, we develop and use a lab-scale setup to test the accuracy of our proposed visual tracking framework for active tracking applications. 
In this paper, we represent the state of the target in the form of the pixel position. Therefore, the terms \lq visual tracking algorithm\rq and \lq target state estimation algorithm\rq are used interchangeably.


With regards to active maneuvering or control of the chaser vehicle, we hypothesize that an efficient controller should cater to the following properties: \textbf{a)} informed by the detection/estimation algorithm;  \textbf{b)} generalize over various target trajectories;  \textbf{c)} compensate for or adapt to the inaccuracies in the target state estimation; and \textbf{d)} computationally efficient for near-real-time application.
Prior research in this area involved designing variants of PID controller and optimization based controllers.
A PID controller was integrated with Lucas Kanade optical flow algorithm to detect and follow ground based objects \cite{bartak2015any}.   
Optimization based control strategies have been designed using game theoritic concepts \cite{quintero2014vision}. A novel UAV path planning algorithm based on elliptical tangent model was proposed in \cite{liu2018novel}. 
A trajectory optimization method was proposed in \cite{wang2021visibility}, with the assumption of target following a smooth trajectory.
These conventional control methods do not meet one or more of the properties mentioned earlier. In that, PID based linear controllers may not be able to handle the non-linearities while the non-linear controllers may not be compute-efficient and generalizable to various target trajectories. In this paper, we propose a neuro-controller to serve as the active maneuver planning (namely velocity control) method, which is trained using policy gradient techniques over an AirSim \cite{shah2018airsim} based simulation environment. 
For the rest of the paper, we will use the term policy instead of neuro-controller to represent the Deep Neural Network (DNN) based controller. 

The key contributions of this paper can thus be summarized as follows:
\begin{enumerate}
\item Develop a robust yet computationally efficient target state estimation framework that innovatively combines KCF and YOLO for target state estimation. 
\item Construct and use a lab-scale setup to test the computational complexity and accuracy of target state estimation algorithm.
\item Express the maneuver (velocity) control for active tracking as a novel Markov Decision Process (MDP) informed by the target state estimation, and design a fast neuro-controller by applying policy gradient (RL) approach over this MDP. 
\item Develop and use an AirSim-based simulation environment to train and test the policy for active tracking, with comparisons to a classical control baseline. 
\end{enumerate}

The remainder of the paper is organized as follows:
In Section \ref{sec:method}, we explain our target state estimation algorithm and the MDP formulation. Subsequently, Sec.~\ref{sec:expts-sim} gives the training framework and the numerical experiments. Section \ref{sec:val-target-state} describes the validation of target state estimation algorithm. Finally, we present our conclusion and future work in Sec.~\ref{sec:conc}.


\vspace{-0.3cm}
\section{Method} \label{sec:method}
\begin{figure}
    \centering
    \includegraphics[width=0.95\linewidth]{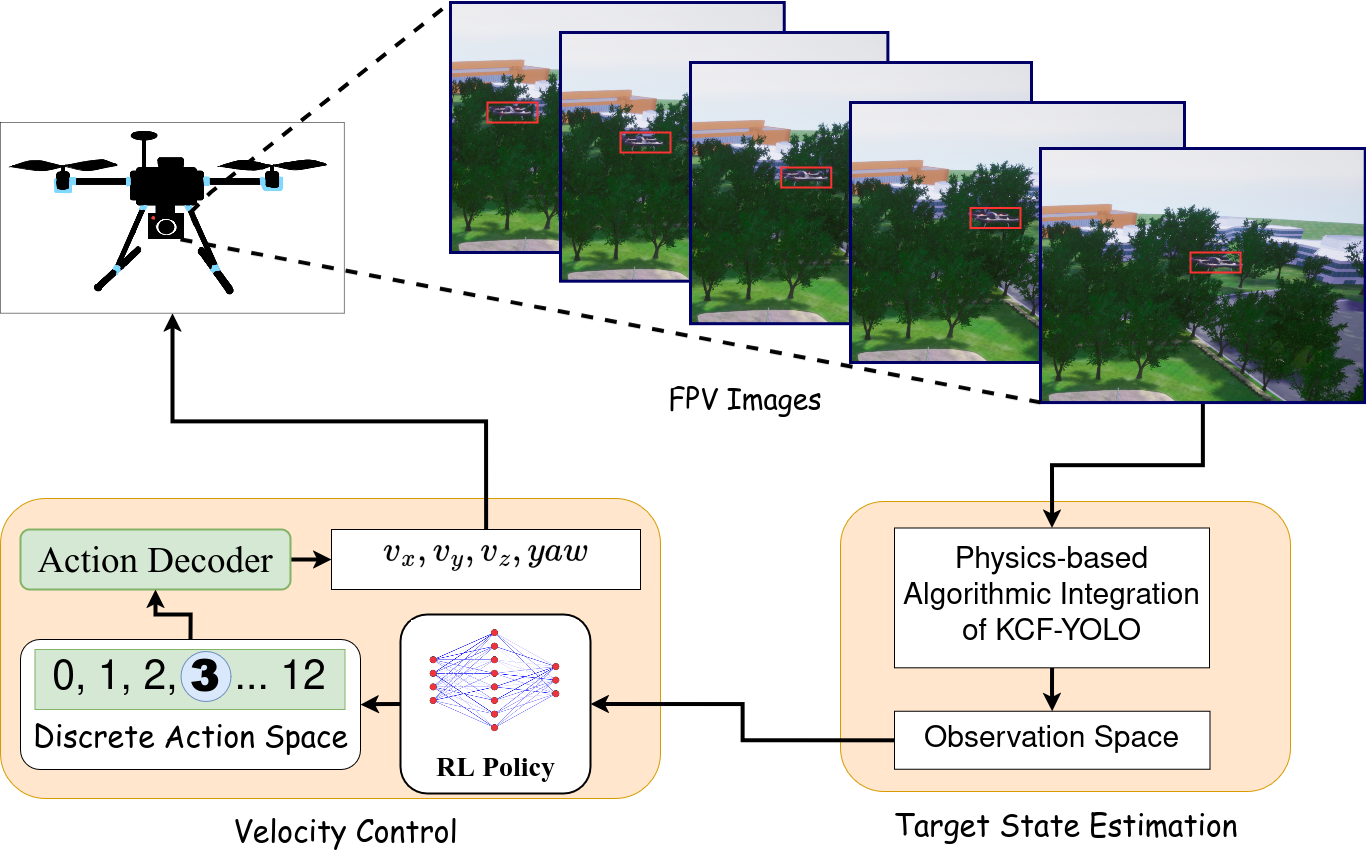}
    \caption{Vision-Based Active Tracking Framework}
    \vspace{-0.5cm}
    \label{fig:overall-framework}
\end{figure}


The entire pipeline of vision-based active tracking using the proposed framework is shown in Fig.~\ref{fig:overall-framework}. 
The images captured by the First Person View (FPV) camera on-board the chaser are first processed by the target state estimation algorithm. The bounding box around the target given by the algorithm forms the input to the policy. The policy outputs linear velocities and yaw-rate for the chaser. The on-board controller (e.g. pixhawk) converts the velocities into motor velocities and executes them. 
In the remainder of this section, we first give the preliminaries required for the target state estimation algorithm. Following that, our target state estimation algorithm is explained. Subsequently, we explain the state space, action space and reward formulation developed to train the policy for velocity control of the chaser. 

\subsection{Preliminaries}
\subsubsection{Kernelized Correlation Filter (KCF)} \label{sec:kcf-prem}
KCF, first proposed in \cite{henriques2014high} is coorelation-filter based visual-tracking approach.
In the first image, the user has to provide the ground truth in the form of bounding box coordinates around the object of interest. The correlation filter $w$ is learnt by fitting a gaussian peak over the bounding box with the peak at the center and solving a regression problem formulation in kernel space. The bounding box becomes the search window for the next frames. A response map is obtained by the correlation of the image patch in the search window and the weight matrix.
The peak of the response map is considered to be the target position. The search window is then updated such that the target position is in the center of the window. 

\subsubsection{Average Peak to Correlation Energy (APCE)} \label{sec:apce}
Average Peak to Correlation Energy is a confidence measure, used in the visual object tracking algorithms to assess the quality of the tracking. The output of the correlation operation of the image $I$ with the correlation filter $w$, given by ($I \circ w$) is a response map ($R$). The APCE is a measure of strength of the peak value in comparison to the other values in the response map and can be computed using Eqn.~\ref{eqn:apce}.  

\begin{equation} \label{eqn:apce}
APCE = \frac{|R_{max} - R_{min}|^2}{\frac{1}{W \times H} \sum_{w=1}^W \sum_{h=1}^H (R(w,h) - R_{min})^2}
\end{equation}
$W, H$ is the width and height of the search window. 

\subsection{Target State Estimation: YOLO-KCF Integration} \label{subsec:yolo-kcf-int}
\begin{figure*}[h]
    \centering
    \includegraphics[width=\linewidth]{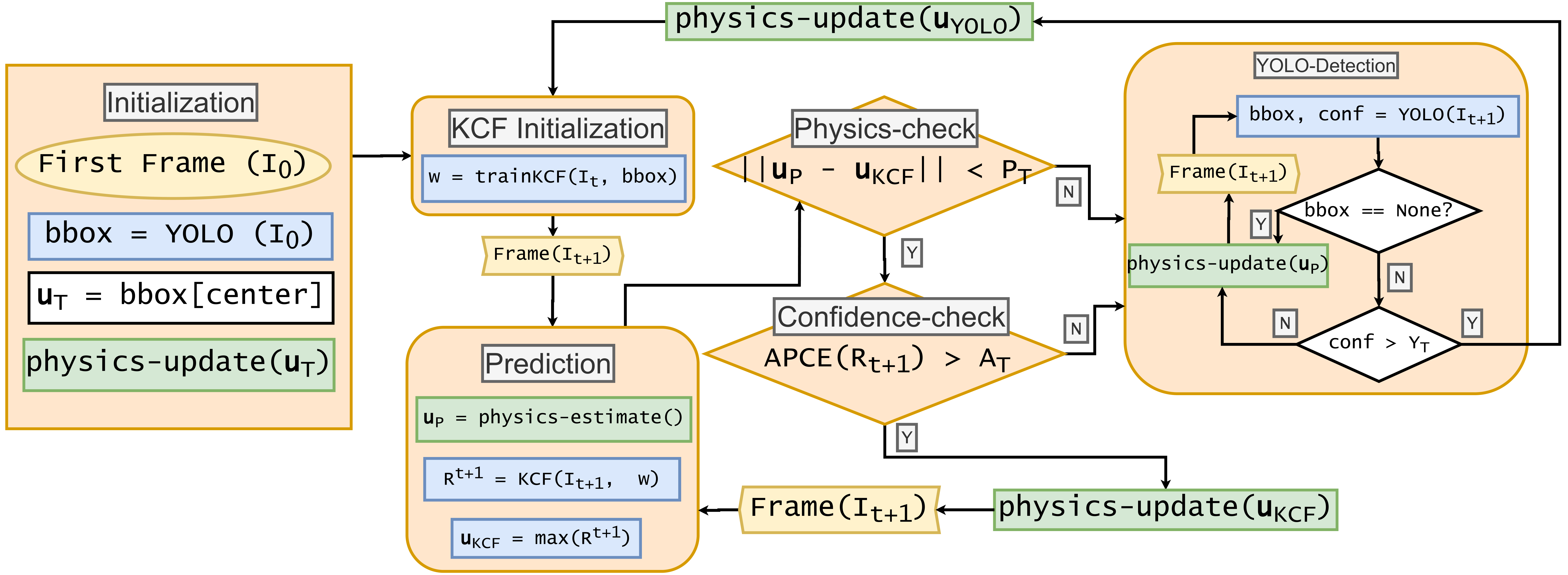}
    \caption{Proposed target state estimation algorithm.
    $\mathbf{u_T}$ is the target pixel position. $P_T$ is the pre-set threshold for physics check, and $A_T$ is the pre-set threshold for APCE. 
    \texttt{trainKCF} refers to obtaining correlation filter $w$ by solving regression problem (Sec.~\ref{sec:kcf-prem}). \texttt{KCF} refers to infering target pixel position $\mathbf{u_T}$ in the frame $I^t$ by using $w$. 
    $\mathbf{R^t}$ represents the response region given by KCF at time for the image frame $I_t$. Physics update corresponds to updating the target velocity estimation. $bbox$ refers to the bounding box and $conf$ is the confidence of prediction.}
    \label{fig:target-est}
\end{figure*}

The proposed state estimation framework uses two perception algorithms: YOLO and KCF. 
We use KCF as the main visual tracking algorithm, while YOLO is invoked for initialization, no-detection, and re-detection. Figure~\ref{fig:target-est} shows the flowchart of the complete algorithm. At the start of the mission, it is assumed that the target is in the FoV of the chaser. Therefore, we use YOLO to detect the target in the first frame and obtain the bounding box (later becomes the search window) required to train the correlation filter for KCF. 
This is shown in the \texttt{Initialization} block (first block on the left) in Fig.~\ref{fig:target-est}. After initialization, KCF is used to track the target in the successive frames. However, KCF outputs the target position in the search window. Therefore, the absence of a target in the search window would make the KCF output invalid. In order to validate the KCF output, we use two criteria. 

First, we validate the KCF output ($\mathbf{u_{KCF}^t}$) for the image frame ($I_t$) using the velocity estimations of the target. 
The position data of the last two frames is used to compute the velocity estimate. This velocity estimate is used to derive the estimate of the target position at timestep $t$ ($\mathbf{u_{P}^t}$). If the norm of the difference between $\mathbf{u_{KCF}^t}$ and $\mathbf{u_{P}^t}$ is less than the threshold, we use our second criterion to validate the KCF output. We use APCE (Sec .~\ref {sec:apce}) to compute the confidence of the KCF prediction. If the confidence is above the threshold, the KCF prediction ($\mathbf{u_{KCF}^t}$) is considered valid and is used to update the velocity estimations (\texttt{physics-update} in Fig.~\ref{fig:target-est}). Further, KCF will be used to compute the target position in the next frame ($I^{t+1}$). If any of the above criteria is not met, YOLO is invoked for detection (\texttt{YOLO-Detection} block in Fig.~\ref{fig:target-est}). YOLO outputs a bounding box and the confidence of the detection. If the confidence threshold is met, the output provided by YOLO is used to update the velocity estimates and obtain the new KCF filter. Otherwise, we consider the YOLO detection as invalid (no-detection) and use the previous position estimates to update the velocity estimations. In that case, we continue using YOLO for the next frame until a valid detection is obtained. Once a valid detection is obtained, we re-initialize KCF and learn a new filter which will then be used for successive frames. 

\subsection{MDP formulation}
Based on the target state information provided by the state estimation algorithm, the policy needs to output the control velocities for the chaser. 
Therefore, the output of the target state estimation algorithm is included in the input (observation space) to the policy, 
while the output of the policy (action space) constitutes the linear velocities and the yaw rate.
Next, we explain in detail the observation space, action space and reward design.  

\subsubsection{Observation Space}
The required velocity of the chaser should be intuitively decided by the relative target position, current velocity and the yaw-angle of the chaser. Therefore, our observation tuple $ \mathbf{o_t} = [x_{\text{min}}, y_{\text{min}}, x_{\text{max}}, y_{\text{max}}, d, v_x, v_y, v_z, \theta]$ consists of the bounding box coordinates of the target ($[x_{\text{min}}, y_{\text{min}}, x_{\text{max}}, y_{\text{max}}]$), euclidean distance between the chaser and the target ($d$), linear velocities of the chaser ($v_x, v_y, v_z$) and the yaw angle of the chaser ($\theta$). The target trajectory can be better captured by looking at a stack of the target positions. Therefore, past 5 such observation tuples are included in the observation space $\mathcal{O} = \mathbf{\{o_t\}}_{t-5}^{t}$. 
\subsubsection{Action Space}
Velocity control requires policy to output desired velocities of the chaser. We choose linear velocities $(v_x, v_y, v_z)$ and yaw rate ($\dot{\theta}$) -- as the control actions. The linear velocities enable the chaser to adjust its position in the 3D world and yaw rate is used to adjust the heading so that the target is in the FoV. 
We propose using a discrete actions space. Table \ref{tab:action-space} shows the actions and the corresponding descriptions for each action.
We have 13 possible actions. 
The forward and backward movements are enabled through actions 0 and 1. To control the yaw rate, we use actions 2 and 3. Action 4 enables chaser to move forward and turn left, while action 5 enables the chaser to move forward and turn right. Action 6 and 7 move the chaser backward and turn left or right. The next four actions $(8, 9, 10, 11)$ move the chaser to one of the four quadrants. The last action ($12$) makes the chaser hover at its current position.
A discrete action space in essence indicates the direction of travel for the chaser. The actions can then be scaled to according to the environment, chaser velocity bounds and the estimated target velocity.

\begin{table}[]
\footnotesize
    \centering
    \begin{tabular}{|c|c|c|c|c|c|}
    \hline
         Action & $v_x$ & $v_y$ & $v_z$ & yaw rate ($\dot{\theta}$) & Description  \\
         \hline
         0 & 1 & 0 & 0 & 0 & Forward \\
          \hline
          1 & -1 & 0 & 0 & 0 & Backward \\
         \hline
         2 & 0 & 0 & 0 & 30 & Turn Left \\
         \hline
         3 & 0 & 0 & 0 & -30 & Turn Right \\
         \hline
         4 & 1 & 0 & 0 & 30 & Forward \& Turn-Left \\
         \hline
         5 & 1 & 0 & 0 & -30 & Forward \& Turn-Right \\
         \hline
         6 & -1 & 0 & 0 & 30  & Backward \& Turn-Left \\
         \hline
         7 & -1 & 0 & 0 & -30 & Backward \& Turn-Right \\
         \hline
         8 & 0 & 1 & -1 & 0 & Top right \\
         \hline
         9 & 0 & -1 & -1 & 0 & Top Left \\
         \hline
         10 & 0 & 1 & 1 & 0 & Bottom Right \\
         \hline
         11 & 0 & -1 & 1 & 0 & Bottom Left \\
         \hline
         12 & 0 & 0 & 0 & 0 & Hover\\
         \hline
    \end{tabular}
    \caption{Discrete Actions}
    \vspace{-0.5cm}
    \label{tab:action-space}
\end{table}

\subsubsection{Reward function}

In order to track the target for longer time-steps, target should be in the center of FoV in the future time-steps and at the desired distance from the chaser. The reward function was designed to train the policy to learn the above requirements. It consitutes three terms: 1) Alignment reward ($R_a$), given by Eqn.~\ref{eqn:ra} encourages the policy to keep the target in the center of the future frame, 2) Distance reward ($R_t$), given by Eqn.~\ref{eqn:rt} is designed to encourage the policy to maintain the desired distance. 3) Through experiments we observed that the reward function $R = R_a + R_t$ led to premature convergence. Therefore, we include another reward term to encourage target tracking for longer time-steps ($R_c$). $R_c$ given by Eqn.~\ref{eqn:rc} rewards the policy for more number of continous detections. 

\begin{equation}\label{eqn:ra}
     R_{\text{a}} = 
     \exp\left(-\alpha_1 \cdot \frac{\mathbf{u}_{T}}{\sqrt{w^2 + h^2}}\right)
\end{equation}
where $\mathbf{u}_{T}$ = distance of the target from the image center, $w, h$ is the image resolution. 

\begin{equation}\label{eqn:rt}
    R_{\text{t}} = \exp\left(-\alpha_2 \cdot |d - d^*|\right)
\end{equation}
where $d$ = current distance from the target, $d^*$ = desired target distance between the target and the chaser
 
\begin{equation}\label{eqn:rc}
    R_{\text{c}} = \frac{1}{\exp\left(1 - 2x\right)}
\end{equation}
where x = $\frac{n}{n_{max}}$, $n$ = continuous detection steps, $n_{max}$ = maximum continuous detection steps (maximum episode length)

\section{Training and Numerical Experiments}    \label{sec:expts-sim}
The goal of this work is to train a policy (neuro-controller) which can generalize efficiently to the complex target maneuvers. 
Therefore, we consider the target assuming a sine curve maneuver, where the frequency and amplitude are sampled from a Gaussian distribution. 
The policy should also be able to account for the background variations leading to missed detections. 
Therefore we use AirSim simulator, which provides photo-realistic rendering of the environment. In addition, uncertainties in UAV movements, wind effects are also included in AirSim. An open-source, pre-compiled AirSim environment Africa Savannah (Figure \ref{subfig:sim}) is used to train the policy. Africa Savannah consists of a dense canopy of trees and bushes, making it suitable to a real world application of vision based active tracking of a flying aerial target. 
\subsection{Training}
We use a UAV to represent the flying aerial target. Therefore, we spawn two UAVs in AirSim, a chaser and a target. The chaser is equipped with a camera and the visual data from the camera can be accessed using AirSim API.
We use a pre-trained YOLOv11 model for object detection in our AirSim based policy training pipeline. The model was trained on a set of images available online and data collected and labeled manually.  We do not use the proposed target state estimation framework in the policy training pipeline because computational cost is not a concern during training on workstations. 

\begin{figure}[t]
    \centering
    \includegraphics[width=0.95\linewidth]{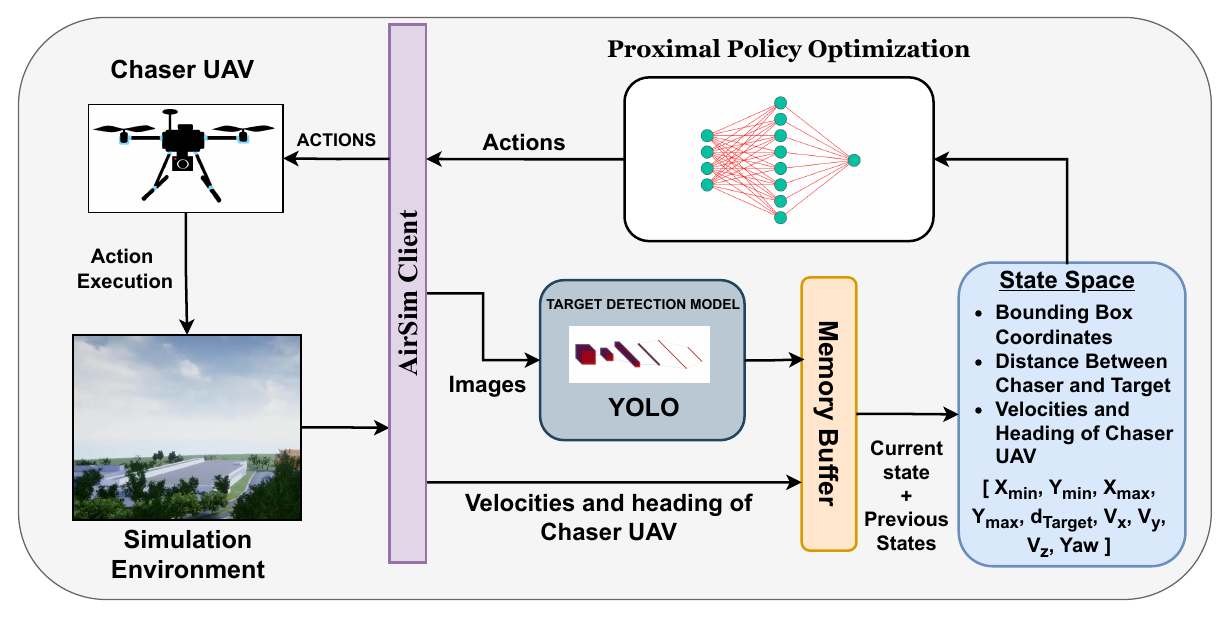}
    \caption{Proposed Training framework}
    \vspace{-0.5cm}
    \label{fig:train-rl}
\end{figure}

At the start of each episode, the UAVs are spawned at height of 50 m, with the target in the FoV of the chaser. The target UAV is given velocity control through AirSim API such that it follows a sine curve ($A\sin 2\pi ft$), either on the $XY$ plane or the $YZ$ plane. For the target maneuver on $XY$ plane, $v_x = 0.5, \hspace{1mm} v_y = A\sin 2\pi ft, \hspace{1mm} v_z = 0$ and for $YZ$ plane, $v_x = 0.5, \hspace{1mm} v_y = 0, v_z = A\sin 2\pi ft$; where $A \sim \mathcal{N}(5, 1)$, $f \sim \mathcal{N}(0.1, 0.05)$. 
At each step, an image from the onboard camera of the chaser is retrieved using the AirSim API, and passed through YOLOv11 which gives bounding box coordinates ($[x_{\text{min}}, y_{\text{min}}, x_{\text{max}}, y_{\text{max}}]$).  
The confidence threshold for YOLOv11 was set to $75$\%. The detections which do not meet the threshold are considered invalid. The bounding box coordinates is set to $[0, 0, 0, 0]$ in case of an invalid detection or no-detection. 
The current velocity of the chaser ($v_x, v_y, v_z$), and its yaw-angle ($\theta$) is obtained using the AirSim API. 
The bounding box coordinates along with the current velocity forms an observation tuple. It gets saved to a memory buffer and last 5 such observation tuples make the state space which is given to the policy.
The output of the policy is a discrete action, and the corresponding velocities from the Table \ref{tab:action-space} are executed by the chaser.  This training pipeline is presented pictorially in Fig.~\ref{fig:train-rl}.
The termination of an episode is based on two criteria 1) Maximum-Steps: The episode is terminated if the chaser UAV is able to track the target for the maximum number of timesteps. This is set to 500 for training. 2) Maximum-Continuous-No-Detection-Steps: If the target is not detected or the detection is invalid continuously for a number of steps, that would imply the chaser has probably lost the target and it is not in the vicinity. Therefore, the episode should be terminated. This is set to 25. 



\begin{figure}[t]
    \centering
    \begin{minipage}{\linewidth}
        \centering
        \includegraphics[width=0.6\linewidth]{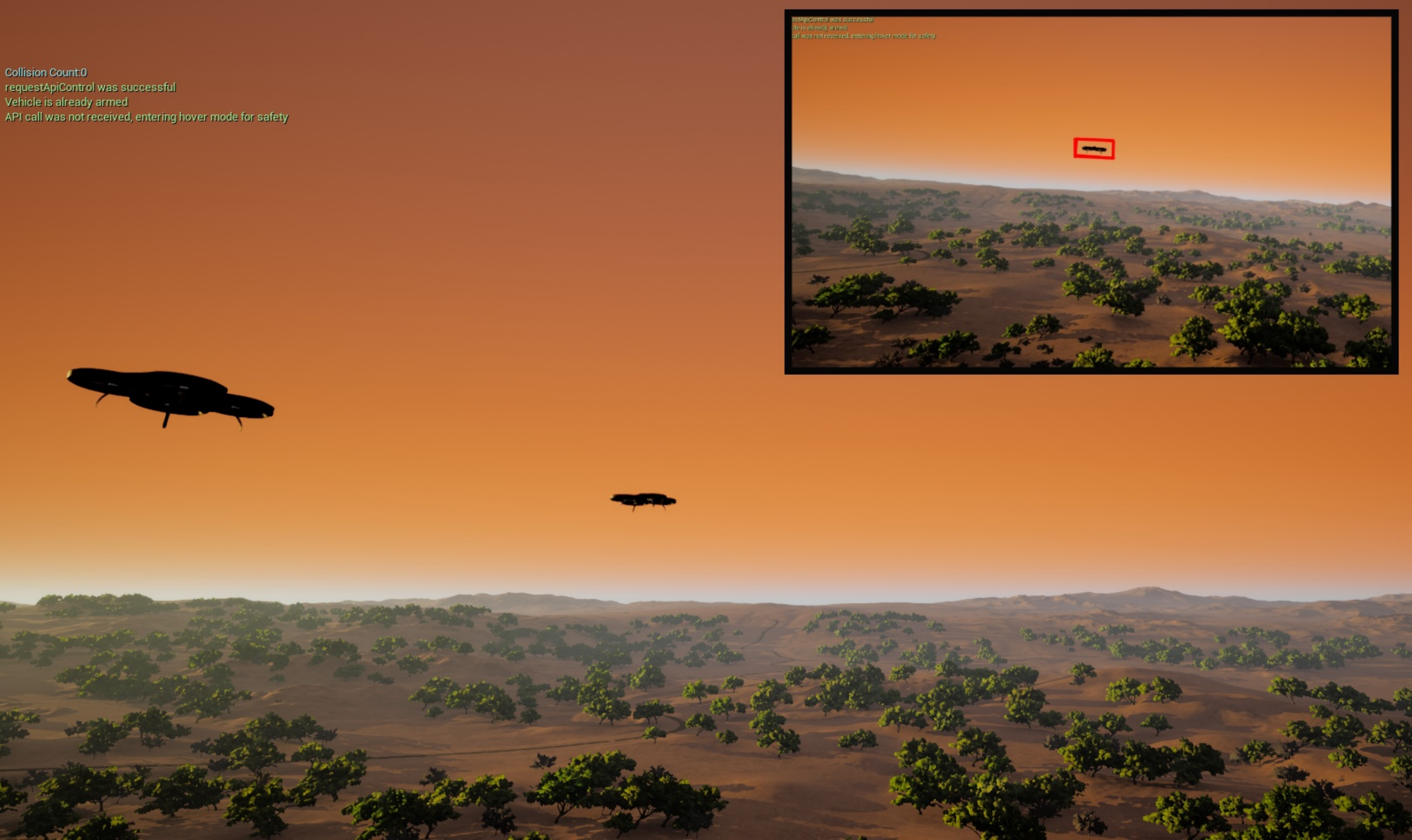}
        \subcaption{AirSim Simulation}
        \label{subfig:sim}
    \end{minipage}

    \vspace{0.5em} 

    \begin{minipage}{0.48\linewidth}
        \centering
        \includegraphics[width=\linewidth]{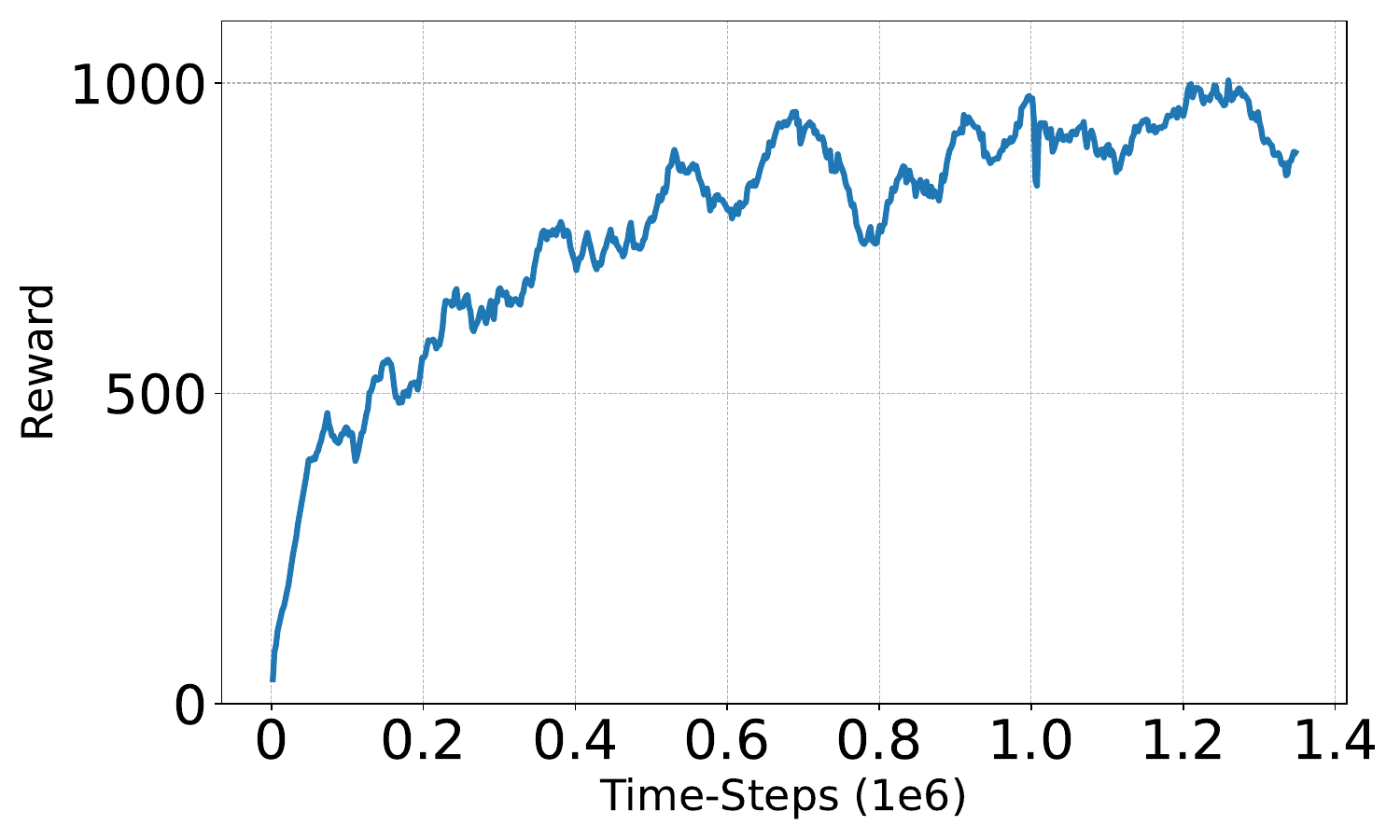}
        \subcaption{Reward Plot}
        \label{subfig:reward}
    \end{minipage}
    \hfill
    \begin{minipage}{0.48\linewidth}
        \centering
        \includegraphics[width=\linewidth]{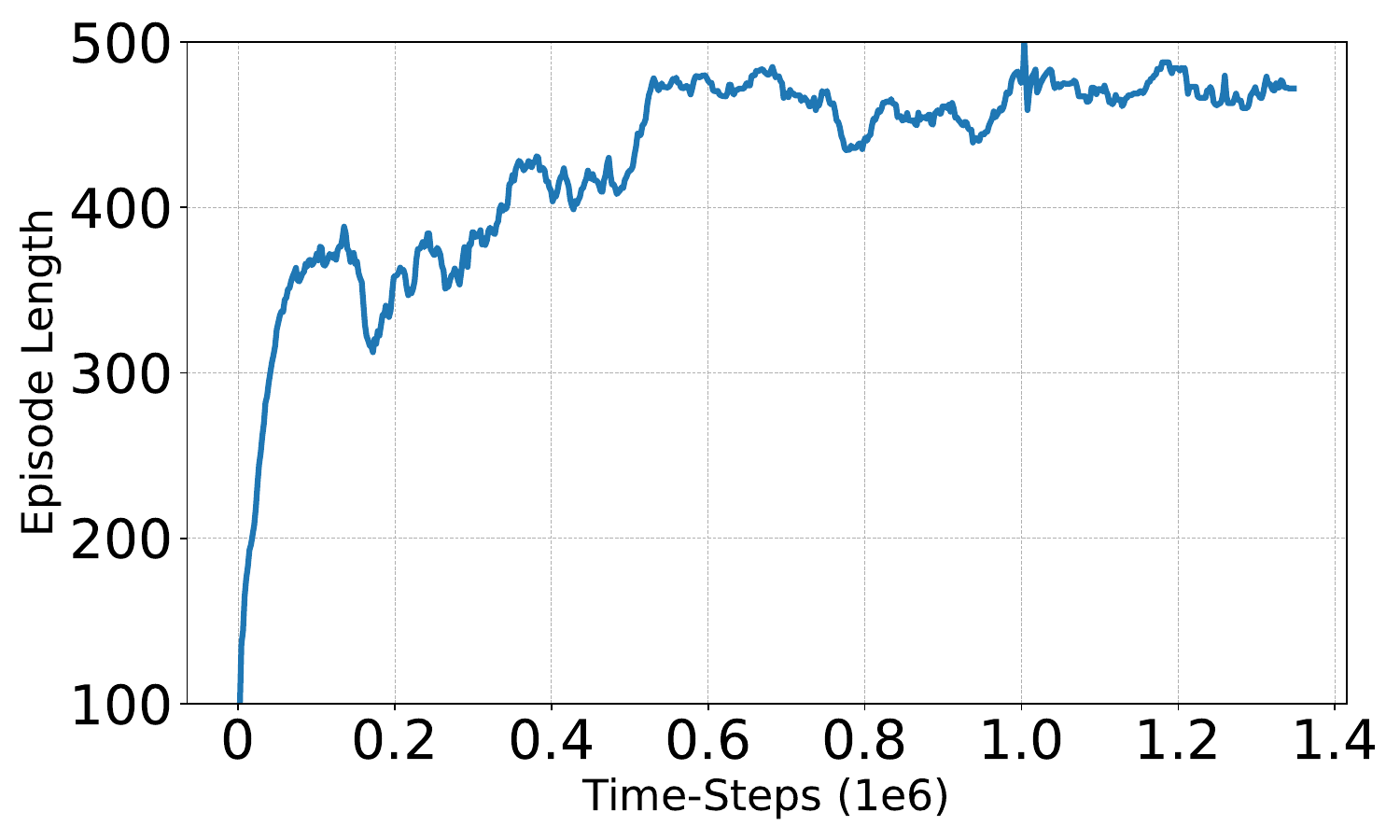}
        \subcaption{Episode Length}
        \label{subfig:length}
    \end{minipage}

    \caption{Training plots from simulation in AirSim}
    \label{fig:reward-length}
\end{figure}

\begin{figure}

\begin{minipage}[h]{0.48\linewidth}
\includegraphics[width=0.85\linewidth]{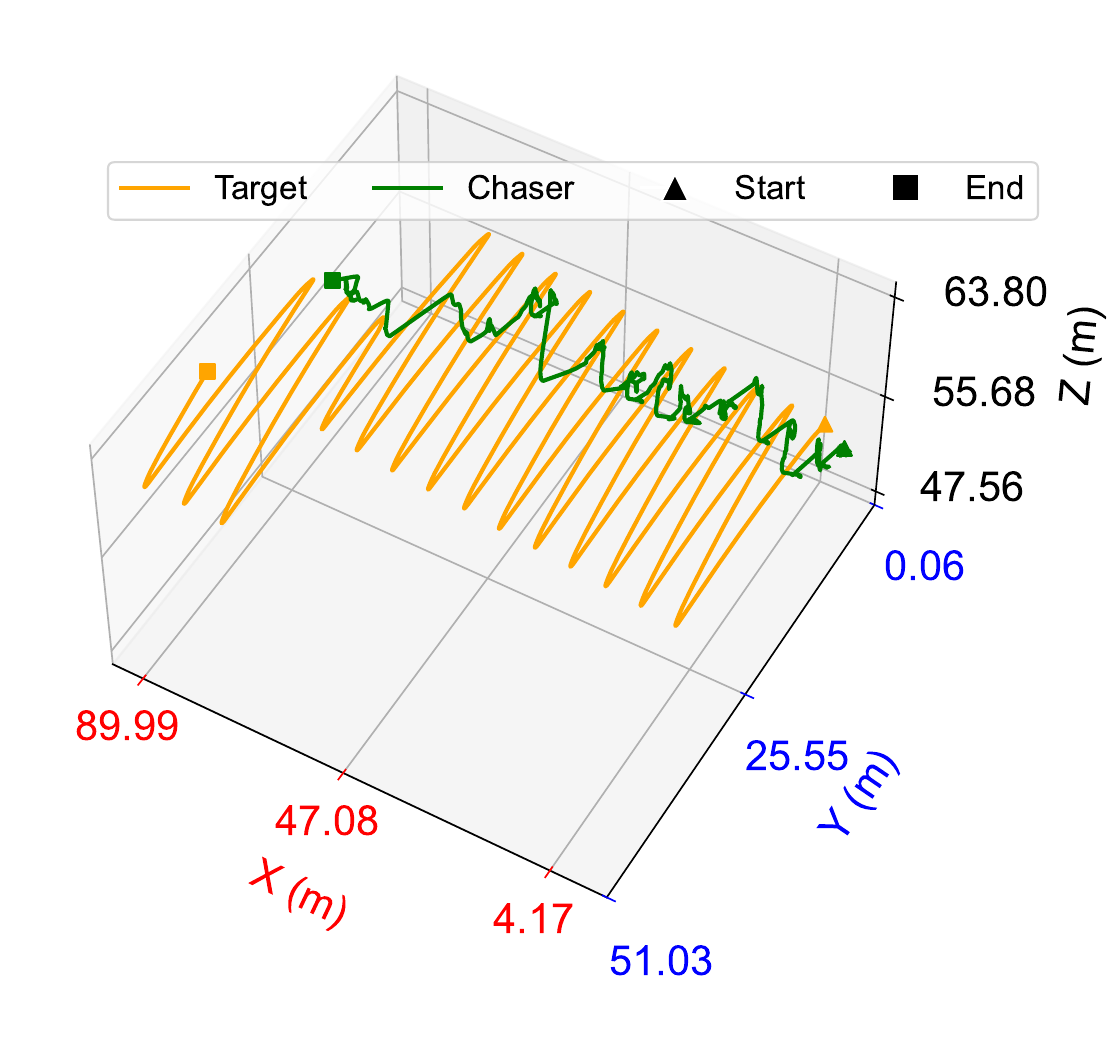}    
\subcaption{} \label{subfig:ha}
\end{minipage}
\begin{minipage}[h]{0.48\linewidth}
\includegraphics[width=0.85\linewidth]{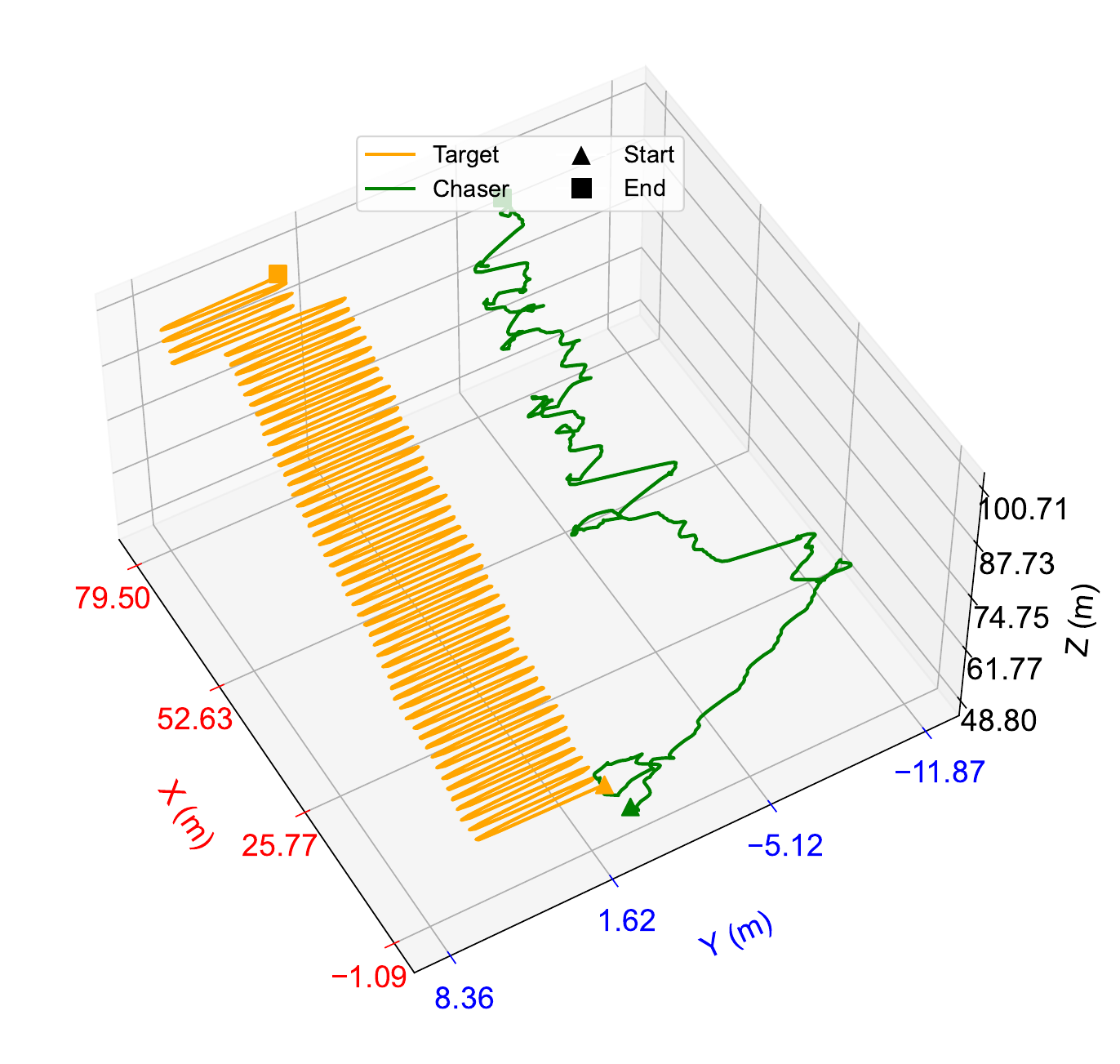}    
\subcaption{} \label{subfig:half}
\end{minipage}


\caption{Target and Chaser trajectories from simulation experiments}
\vspace{-0.5cm}
\label{fig:trajectories}
\end{figure}

We have used the standard Deep Reinforcement Learning (DRL) library stable-baselines3 \cite{stable-baselines3} to train our policy using PPO. The learning rate is set to $1 \times 10^{-4}$, batch size to $256$, and Generalized Advantage Estimation (GAE) set to 0.95. 
Figure \ref{subfig:reward}, \ref{subfig:length} shows the mean reward and episode length for $1.5 \times 10^6$ timesteps of training. There is a high correlation between length of the episode and reward which shows that increasing rewards lead to better episode length.   

The simulation training experiments were conducted on a high-performance Dell Precision 7920 Tower equipped with an Intel Xeon Gold 6148 CPU with 80 cores at 2.40 GHz, 192 GB of memory, and an NVIDIA GeForce RTX 4060 Ti GPU with 16 GB of VRAM.

\subsection{Numerical Experiments}

During training, the target was given a sine maneuver where the Amplitude and frequency were drawn from a normal distribution. Therefore, through the numerical experiments performed in Airsim, we aim to answer the following questions: 1) How well does the policy generalize to the sine maneuver, with the parameters drawn from the same distribution?  2) Can the policy generalize to the sine maneuver with the parameters drawn from the Gaussian distribution with other parameters? 


As a baseline, we use a basic implementation of Image-Based Visual Servoing (IBVS) control mechanism. 
IBVS aims to control the chaser so that the target is in the center of the image frame ($c_x, c_y$) and at the desired distance ($d^{*}$). Therefore, error is given by $e_x = u_t - c_x, e_y = v_t - c_y, e_z = d - d^{*}$ where $u_t, v_t$ are the target positions in the pixels as estimated by the target estimation algorithm.  The linear velocities are given by: $v_x = k_u.e_x, v_y = k_v.e_y, v_z = k_z.e_z$. The error for $yaw\_rate$ is calculated as $e_{\psi} = (u_t - c_x)/d$. The yaw rate is then given by $k_{\psi}.e_{\psi}$. The terms $k_u, k_v, k_z, k_{\psi}$ are gains which need to be tuned. In the current implementation, we tune them for the target following a sine curve with $A = 5$ \& $f=0.1$. We call this IBVS control scheme as PID from now onwards. 
Note that PID will work well only if the target state estimation algorithm works well. In our case, we use YOLO for target state estimation in simulation. Therefore, with PID the target can be tracked only if YOLO prediction is accurate. 


\begin{figure}[h]
  \centering
\begin{minipage}[h]{0.45\linewidth}
    \includegraphics[width=\linewidth]{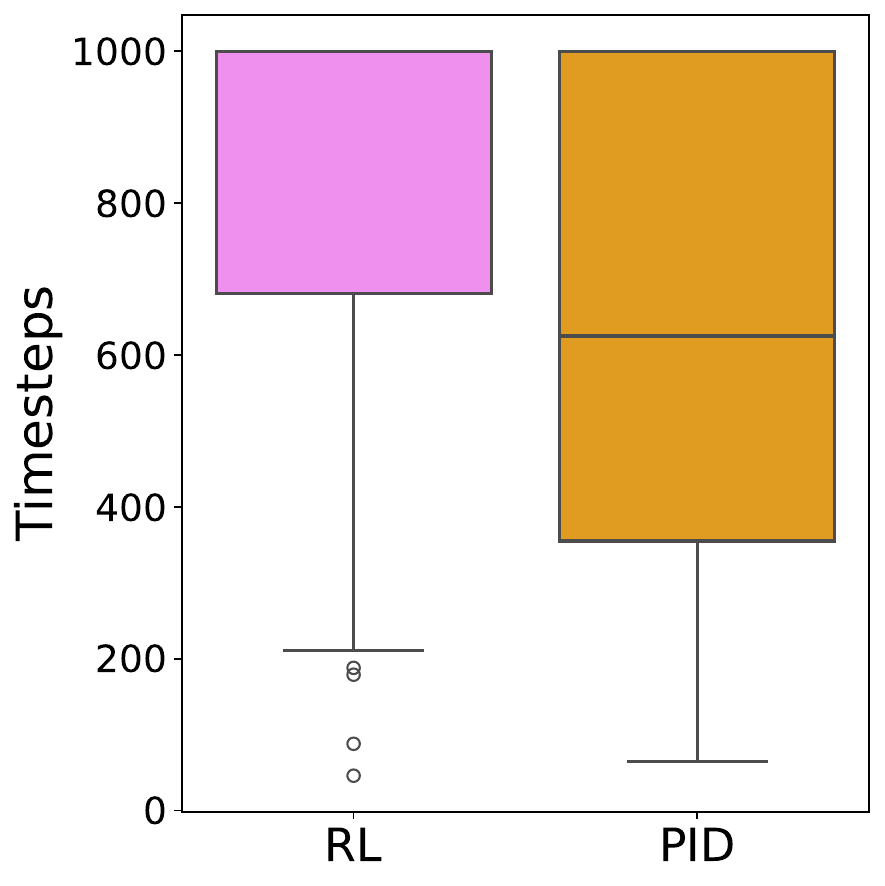}
    \subcaption{Episode Length} \label{subfig:eplen-cs-2-exp-1}
\end{minipage}
\begin{minipage}[h]{0.45\linewidth}
    \includegraphics[width=\linewidth]{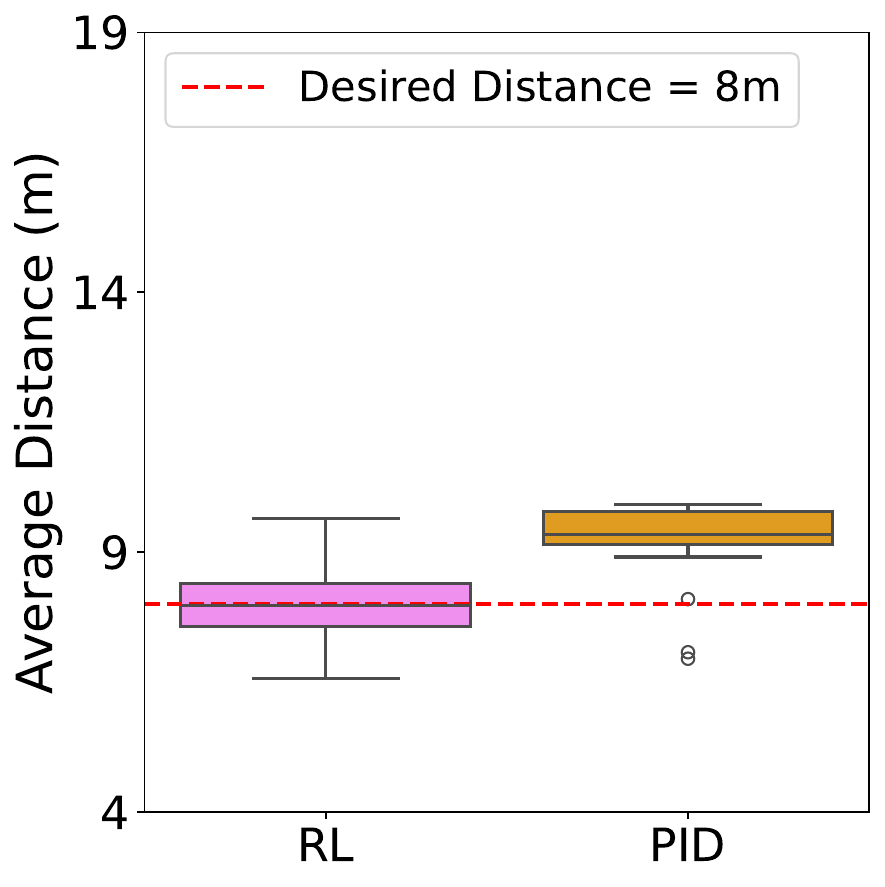}
    \subcaption{Average Distance} \label{subfig:avgdist-cs-2-exp-1}
\end{minipage}
\caption{Controller Evaluation for Case Study 1}
  \label{fig:cs-1}
\end{figure}

\begin{figure}[t]
\begin{minipage}[h]{0.45\linewidth}
    \includegraphics[width=\linewidth]{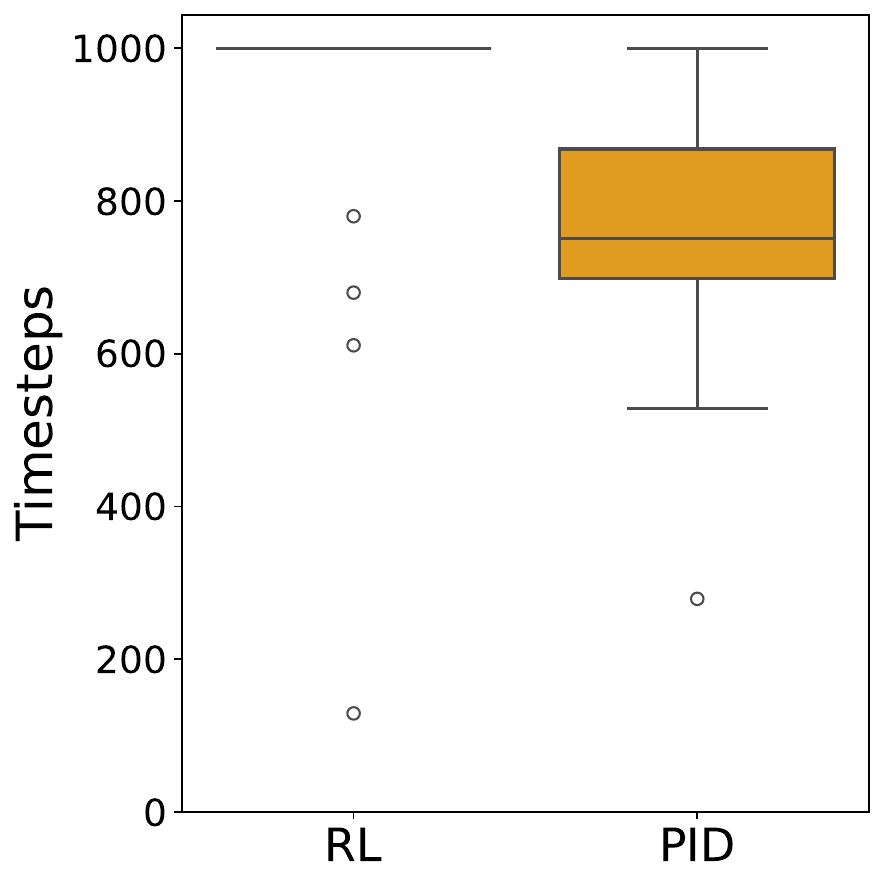}
    \subcaption{Episode Length, Exp-1} \label{subfig:eplen-cs-2-exp-1}
\end{minipage}
\begin{minipage}[h]{0.45\linewidth}
    \includegraphics[width=\linewidth]{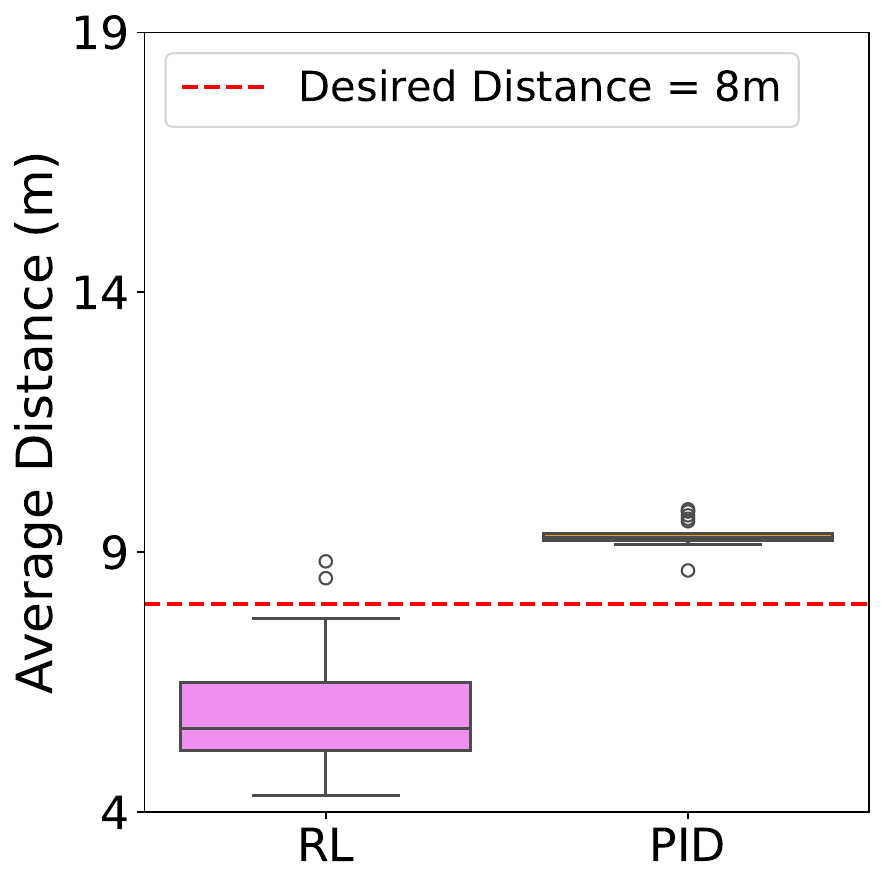}
    \subcaption{Average Distance, Exp-1} \label{subfig:avgdist-cs-2-exp-1}
\end{minipage}

\begin{minipage}[h]{0.45\linewidth}
    \includegraphics[width=\linewidth]{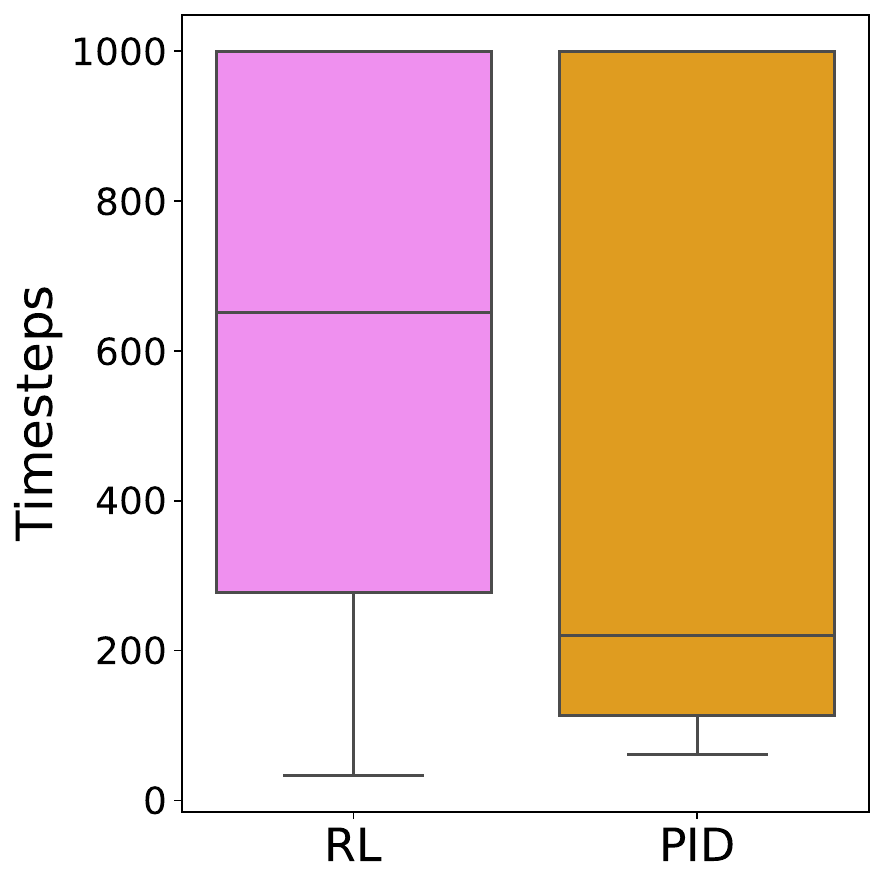}
    \subcaption{Episode Length, Exp-2} \label{subfig:eplen-ha}
\end{minipage}
\begin{minipage}[h]{0.45\linewidth}
    \includegraphics[width=\linewidth]{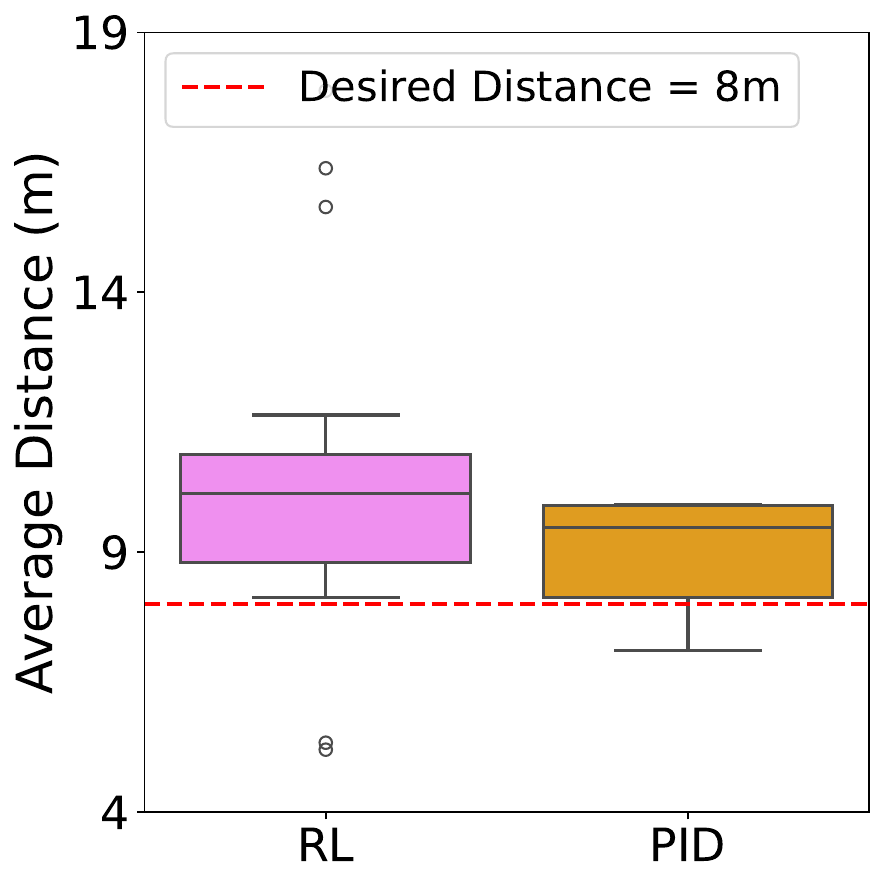}
    \subcaption{Average Distance, Exp-2} \label{subfig:avgdist-ha}
\end{minipage}
\caption{Controller evaluation for case-study 2}
\label{fig:cs-2-exp-1}
\end{figure}


We use two metrics to evaluate the performance of the control algorithms: Episode length and Average distance between the chaser and the target during the episode. Episode length is the total number of time-steps up to which the chaser is able to track the target. The termination criteria of episodes are the same as that used in training: Max-No-Detection (25) and Max-Steps (1000). Note that during training the chaser had to give control inputs for maximum 500 times while we increased that to 1000 for test experiments.  We conducted 2 case-studies and ran 25 episodes for each. In each episode, the target trajectory is sampled randomly in either $XY$ plane ($v_x = 0.5, \hspace{1mm} v_y = A\sin 2\pi ft, \hspace{1mm} v_z = 0$) or $YZ$ plane ($v_x = 0.5, \hspace{1mm}, v_y = 0, v_z = A\sin 2\pi ft$).

\textbf{Case-study 1: Performance evaluation for the observed distribution.}

In the training, the target assumed a sine curve maneuver with the amplitude and frequency sampled from a Gaussian distribution given by 
$A \sim \mathcal{N}(5, 1)$, $f \sim \mathcal{N}(0.1, 0.05)$. Therefore, in this case study we sampled the parameters of the sine curve from the same distribution. Figure \ref{fig:cs-1} shows the box plot for average distance (right) and the number of timesteps (left). The median episode length of the policy is 700 time steps while with PID controller, it is 600. PID also shows higher variance in episode length when compared to the policy. Since the parameters of the target maneuver are sampled from a distribution, each time an episode is initialized the background is expected to change. As noted earlier, PID performance is dependent on the performance of target state estimation algorithm, whereas the policy can also account for the inaccuracies in the state estimation algorithm. 
The median average distance maintained by the policy is equal to the desired distance of $8$ m. The median average distance maintained with PID controller is $9$ m. Thus, the policy was able to maintain desired distance for the experiments conducted in the first case study.

\textbf{Case-study 2: Performance evaluation for a different set of distribution parameters.}

We conduct two experiments for this case-study by varying the distribution of amplitude in each of them. The target is initialized in front of the chaser and since the target performs sine maneuver, the amplitude governs the distance of the target from the line of sight of the chaser. \\
\textit{1) Low Amplitude:} $A \sim \mathcal{N}(1, 1)$, $f \sim \mathcal{N}(0.1, 0.05)$. Since the amplitude mean is lesser than observed in training, the target expected to stay in the FoV for a longer period of time leading to an improved model performance. As shown in Fig.~\ref{subfig:avgdist-cs-2-exp-1}, \ref{subfig:eplen-cs-2-exp-1}, the RL policy is able to track the target over the entire episode duration for all the episodes. The average distance maintained by the chaser is also close to the desired distance. While the performance of both the RL and PID shows improvement, RL still outperforms PID in terms of episode length and average distance. \\
\textit{2) High Amplitude:} $A \sim \mathcal{N}(10, 1)$, $f \sim \mathcal{N}(0.1, 0.05)$. In this experiment, we increase the mean amplitude of the sine curve by $5$. This causes the target to take long and frequent maneuvers in and out of FoV. The wind effects and unstability created by such high speed add to the complexity of tracking mission. This requires the chaser to effectively interpret the target trajectory and take suitable control actions. Figure \ref{subfig:eplen-ha}, \ref{subfig:avgdist-ha} show the results for tracking episode length and average distance maintained during the episode. Both RL and PID has suffered performance in terms of the episode length. However, RL has higher median and lesser variance than PID.   

\begin{figure}[h]
  \centering
  \includegraphics[width=0.95\linewidth]{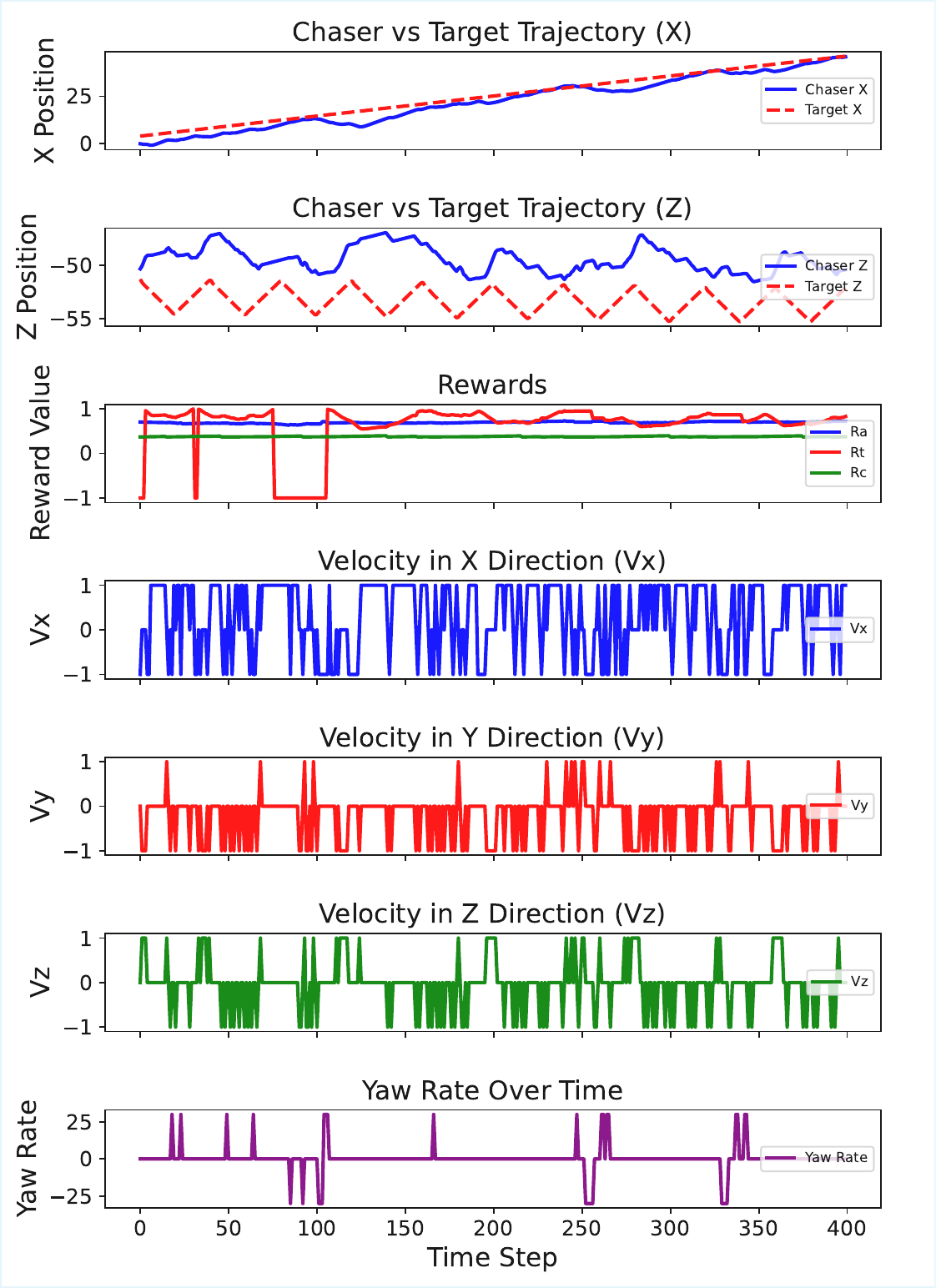}
  \caption{Chaser and Target trajectories, Rewards, and Actions for each time step in an Episode}
  \label{fig:reward-action}
\end{figure}

To get some insights into the performance of the learned policy, we track all the reward metrics ($R_a, R_t, R_c$) along with the action taken at each step for a sample target trajectory. Figure \ref{fig:reward-action} shows the target and chaser trajectories, reward metrics and the actions taken at each time step. The reward curve shows that the tracking reward ($R_t$ represented by red curve in the reward plot) takes negative values when the chaser action leads to an increase in distance between them (such as at 150th timestep) or when the chaser moves too close to the target (100th timestep). The consistency in the alignment reward ($R_a$) and the continuous detection reward ($R_c$) shows that the chaser was able to keep the target in its Field of View over all the timesteps. 
The target is given a constant speed ($0.5$ m/s) in x-direction and the velocity plot shows that the policy gives $v_x = 1$ m/s for more time-steps.

\section{Experimental Validation of Target State Estimation Algorithm} \label{sec:val-target-state}

\begin{figure}[h]
  \centering
  \includegraphics[width=0.85\linewidth]{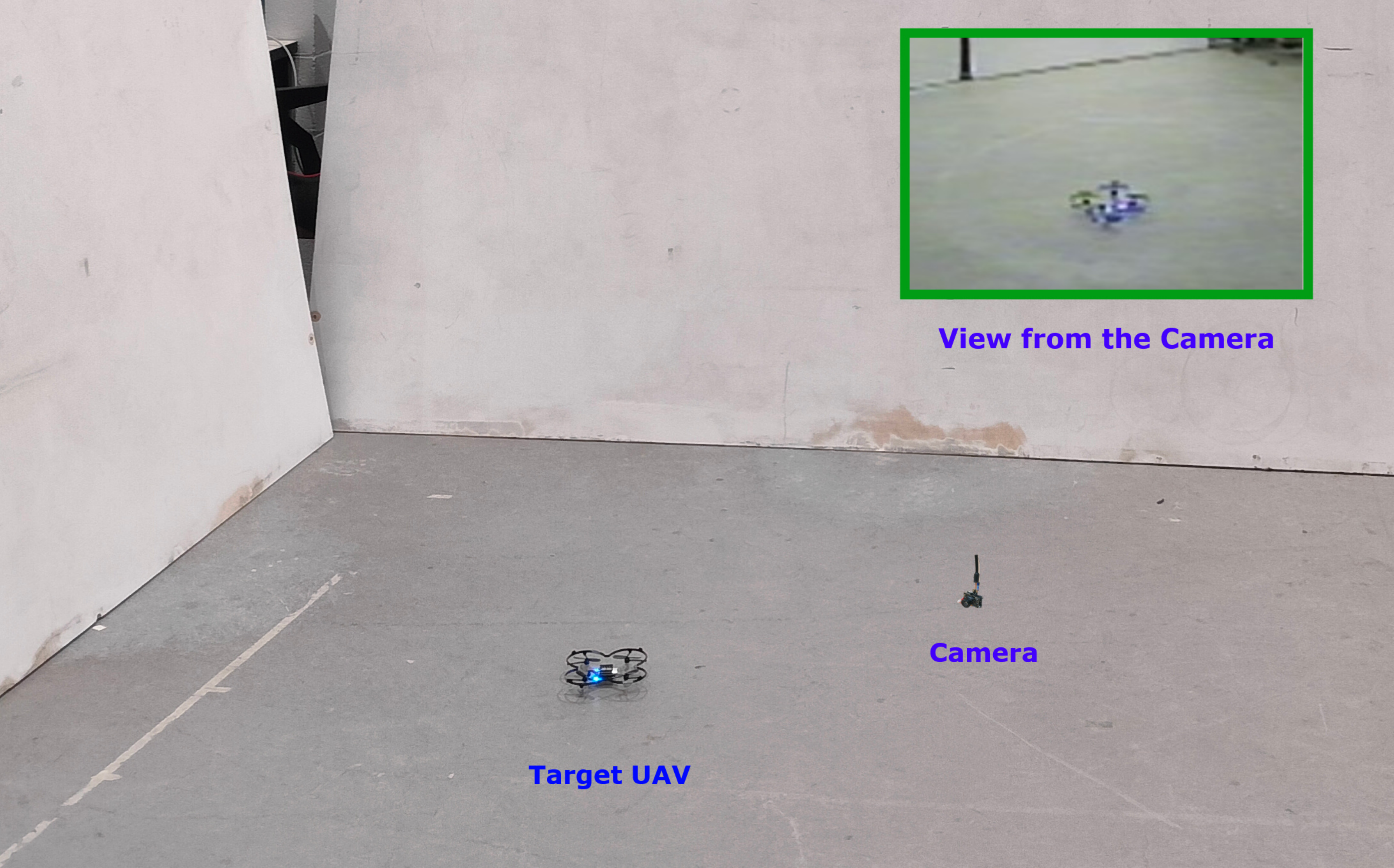}%
  \caption{Experimental setup to create a dataset for validation of target state estimation algorithms in active tracking applications}
  \label{fig:fps-rmse}
\end{figure}

One of the contributions of this paper was to develop a compute-efficient and accurate target state estimation algorithm for active tracking. 
We use a lab-scale setup to create a dataset, which is then used to test the computational complexity and accuracy of the target state estimation algorithm.
The setup, as shown in Fig.~\ref{fig:fps-rmse} has two key components: \textbf{1)} Target object: We use a Crazy flie UAV as the target. \textbf{2)} Camera: We use mini 5.8G FPV 48CH 25mW transmitter VTX-CAM with 600TVL camera. The images captured by the camera are received at the workstation using Skydroid OTG UVC receiver. 
We give a pre-defined trajectory to the target and while the target performs its maneuver, the camera is actively maneuvered by a human with an intention to keep the target in the FoV of the camera. The image feed from the camera is received and recorded on the workstation. 
We created a dataset by giving $4$ types of maneuvers to the target. It included line trajectory, sine trajectory, a circular trajectory and random maneuvers. The dataset had images with the target performing its maneuver over a plain background and over complex backgrounds. The dataset also included images with the target out of the frame, thus effectively testing no-detection and re-detection.

We trained the YOLO model with crazy-flie UAV images. YOLO model outputs confidence along with the bounding box. We set high threshold for the confidence to ensure valid detections. The detections of YOLO model thus serve as ground truth. Recall from Section \ref{subsec:yolo-kcf-int} , our target state estimation framework integrates YOLO and KCF to achieve light-weight and accurate detection. Therefore, we use two metrics to evaluate the performance: FPS (frames per second) and RMSE (root mean squared error). 
FPS indicates the computational efficiency of the target state estimation algorithm and RMSE indicates the accuracy.
We use the newly created dataset to evaluate the performance of each algorithm and the results are shown in Table \ref{tab:perception-results}. 
The RMSE of YOLO is 0 since we use YOLO detections as the ground truth. The min, max and mean FPS of YOLO is 4, 6, 5 respectively. The min, max and mean FPS of our proposed algorithm is 8, 28 and 18. The min, max and mean RMSE of our proposed algorithm is 3, 14 and 7. In contrast, min, max and mean RMSE of KCF is 83, 390 and 208. Therefore, our proposed framework achieves an improvement in FPS as compared to YOLO without compromising on RMSE as compared to kcf. 
We used Alienware M16 laptop with 16th Gen Intel(R) Core(TM) i7 -13700HX, 2100MHz, 16 Cores and 16 GB RAM to conduct evaluations and hence we expect a greater difference in FPS (between YOLO \& YOLO-KCF) when the algorithms are deployed on the processors commonly used in UAVs such as Raspberry pi (Rpi) or Nvidia Jetson Orin Nano. 


\begin{table}[h]
    \centering
    \resizebox{\linewidth}{!}{%
    \begin{tabular}{|c|c|c|c|c|c|c|}
        \hline
        \multirow{2}{*}{Algorithm} & \multicolumn{3}{c|}{FPS} & \multicolumn{3}{c|}{RMSE} \\ \cline{2-7} 
                                 & min & max & mean & min & max & mean \\ \hline
        YOLO                     & 4 & 6 & 5 & 0 & 0 & 0 \\ \hline
        KCF                      & 24 & 52 & 38 & 83.3 & 390.21 & 207.52 \\ \hline
        YOLO-KCF      & 8 & 28 & 18 & \textbf{2.43} & \textbf{14.1} & \textbf{7.52} \\ \hline
    \end{tabular}
    }
    \caption{Evaluation results of the target state estimation algorithm. FPS: frames per second represents compute-efficiency, and RMSE: root mean squared error represents accuracy. YOLO is considered ground truth, so its RMSE is $0$.}
    \label{tab:perception-results}
\end{table}

\vspace{-10mm}
\section{Conclusion} \label{sec:conc}

This paper developed a computationally efficient pipeline for vision based active tracking of a flying target from a chaser UAV. This is accomplished by solving two sub-problems, 1) perception: target detection and state estimation, and 2) maneuver control of the chaser to enable continued detection. Firstly, target detection and state estimation is performed using a combination of correlation-based filter KCF (primary estimator) augmented by confidence triggered calls to the deep learning architecture Yolo (as a corrector). A new dataset was created using a lab-scale setup to successfully validate the target detection algorithm. This perception step of the pipeline executes $3$ times faster than typical standalone implementation of YOLO. 
However, in its current form, the target velocities are estimated in the pixel space based on constant acceleration assumptions, which can lead to uncertainties in position estimate; Extended Kalman Filter (EKF) can be incorporated in the future to capture these uncertainties. 
Compared to the limited preliminary experimental validation provided here, future work could develop and execute a rigorous design of experiments to analyze the compute-efficiency and accuracy of the perception pipeline, the trade-offs involved therein, its effect on the controller and active tracking mission performance. 

Secondly, the problem of active maneuver (velocity and yaw control) of the chaser UAV was performed by a neural network policy trained with RL over an AirSim environment including various target trajectories and background scenes. This policy executes in less than 0.5 ms, and is found to be 42\% better than a PID baseline in terms of the length of time over which continual tracking of target is achieved in unseen test scenarios. 
While the presented control policy was found to be generalizable over unknown target maneuvers, its black-box nature can limit interpretability in general and robustness in edge cases; future work can address both of these concerns by incorporating physics based constraints on the action space and/or reward space. Such extensions, along with consideration of a variety of target types (varying in shape, size and inertial properties), would provide more comprehensive insights into the potential for transitioning this promising learning-aided vision-based air-to-air tracking approach to the field.

\bibliographystyle{ieeetr}
\bibliography{references}

\end{document}